\documentclass[journal,twoside]{IEEEtran}

\usepackage{cite}
\usepackage{graphicx}
\usepackage{psboxit,pstricks}
\usepackage{epsfig}
\usepackage{pmat}
\usepackage{amssymb}
\usepackage{array}
\usepackage{algorithm}
\usepackage{algpseudocode}
\usepackage{color}

\ifCLASSINFOpdf
\else
\fi
\usepackage[cmex10]{amsmath}
\usepackage{array}
\begin{document}
%
\title{3D Morphology Prediction of Progressive Spinal Deformities from Probabilistic Modeling of Discriminant Manifolds}
%
%
%

\author{Samuel~Kadoury,~\IEEEmembership{Member,~IEEE,}
            	William Mandel, Marjolaine Roy-Beaudry, Marie-Lyne Nault and~Stefan~Parent
\thanks{Copyright  $\copyright$ 2016 IEEE. Personal use of this material is permitted. However, permission to use this material for any other purposes must be obtained from the IEEE by sending a request to pubs-permissions@ieee.org.}
\thanks{This work was supported in part by the National Science and Engineering Research Council (NSERC).}
\thanks{S. Kadoury and W. Mandel are with \'Ecole Polytechnique de Montr\'eal, P.O. Box 6079, Succ. Centre-ville, Montr\'eal, Qu\'ebec, Canada, H3C 3A7}%
\thanks{S. Kadoury, M. Roy-Beaudry, M-L. Nault and S. Parent are with the Sainte-Justine Hospital Research Center, 3175 Cote-Sainte-Catherine Rd., Montr\'eal, Qu\'ebec, Canada, H3T 1C5}%
\thanks{Manuscript received September 6, 2016}%
\thanks{Color versions of one or more figures in this paper are available online at
http://ieeexplore.ieee.org.}}

%
%

\markboth{IEEE TRANSACTIONS ON MEDICAL IMAGING}%
{Kadoury \MakeLowercase{\textit{et al.}}: Prediction of Spinal Deformity Progression from Probabilistic Manifolds}
%



\maketitle

\begin{abstract}
We introduce a novel approach for predicting the progression of adolescent idiopathic scoliosis from 3D spine models reconstructed from biplanar X-ray images. Recent progress in machine learning have allowed to improve classification and prognosis rates, but lack a probabilistic framework to measure uncertainty in the data. We propose a discriminative probabilistic manifold embedding where locally linear mappings transform data points from high-dimensional space to corresponding low-dimensional coordinates. A discriminant adjacency matrix is constructed to maximize the separation between progressive and non-progressive groups of patients diagnosed with scoliosis, while minimizing the distance in latent variables belonging to the same class. To predict the evolution of deformation, a baseline reconstruction is projected onto the manifold, from which a spatiotemporal regression model is built from parallel transport curves inferred from neighboring exemplars. Rate of progression is modulated from the spine flexibility and curve magnitude of the 3D spine deformation. The method was tested on 745 reconstructions from 133 subjects using longitudinal 3D reconstructions of the spine, with results demonstrating the discriminatory framework can identify between progressive and non-progressive of scoliotic patients with a classification rate of 81\% and prediction differences of 2.1$^{o}$ in main curve angulation, outperforming other manifold learning methods. Our method achieved a higher prediction accuracy and improved the modeling of spatiotemporal morphological changes in highly deformed spines compared to other learning methods.

\end{abstract}

\begin{IEEEkeywords}
Adolescent Idiopathic Scoliosis, 3D spine reconstruction, Curve progression, Discriminant manifolds, Deformation prediction \end{IEEEkeywords}

%
\IEEEpeerreviewmaketitle

\section{Introduction}
%
%
%
%
\IEEEPARstart{A}{dolescent} idiopathic scoliosis (AIS) is a three-dimensional (3D) deformation of the spine with unknown aetiopathogenesis. For children between 10 and 18 years old, the prevalence of AIS with a principal curvature greater than 10$^{\circ}$ is of 1.34\%. A large scale study demonstrated that close to 40\% of children screened at school and subsequently followed by a clinician are diagnosed with AIS \cite{fong2010meta}. One of the most challenging problems in AIS is the effective prediction of curve progression from a patient's baseline visit, once they are diagnosed with this pathology. In current clinical practice, factors such as patient maturity, both in terms of age and skeletal stage using the Risser sign, menarchal status, curve magnitude and curve location are used to assess a curve's probability for progression. These parameters are often used to establish treatment strategies, such as surgery or orthopedic braces, as well as scheduling follow-up examinations. Methods based on alignment charts were made by \cite{lonstein1984prediction} to link progression incidences with specific types of deformation, however these were primarily proposed to determine the appropriate treatment strategy. Curve progression has become the primary concern for patients and their families as it can cause significant distress from aesthetic and lifestyle perspectives. 


In recent years, spine morphology and in particular 3D morphometric parameters have shown significant promise to assess the link with respect to curve progression. In orthopaedics, 3D reconstructions obtained from diagnostic scans can help orthopedists assess deformations and establish treatment strategies, by providing a personalized model and localize landmarks for deformed inter-vertebral segments. A retrospective evaluation in 3D parameters based on spinal and vertebral morphology was made to classify progressive and non-progressive patients \cite{nault2013three}. More recently, a prospective study was performed to evaluate the differences in 3D morphological spine parameters between both AIS groups using the patient's first visit data \cite{nault2014three}. These prediction systems cluster operator-crafted parameters, directly derived from the 3D spine models. Unfortunately, relying on geometric parameters implicitly requires to determine optimal features which can best represent the true nature of 3D scoliotic spines.

Contrary to explicit parametric models, numerical or statistical methods are capable to describe in a low-dimensional space, the highly dimensional and complex nature of the global geometric 3D reconstruction of the spine, as well as the local variations based on vertebral shapes . Ultimately, 3D spine models could be interpreted implicitly instead of using expert-based features as it was done in previous studies. While wavelet-based compression was used to assess spine curvature \cite{Duong06},  manifold learning performed  on locally linear embeddings was able to reduce the dimensionality of thoracic 3D spine models \cite{Kadoury12}. Non-linear manifolds were explored in many previous works using  Gaussian distribution with probability features \cite{Lawrence05} and with spectral signatures \cite{Kanaujia07}. Laplacian Eigenmaps or locally linear embedding \cite{Roweis00} are some of the learning algorithms which maps high-dimensional features which is assumed to belong to a non-linear domain, onto an underlying sub-space of reduced dimensionality which maintains the global structure. This is achieved with the conservation of local relationships of similar object geometries. Unfortunately, these dimensionality reduction algorithms based on local estimation are prone to be affected by the out of sample problem and sensitive to samples which map outside the normal distribution of the observed data \cite{van2009dimensionality}. 

\begin{figure}
\begin{center}
\begin{minipage}[b]{0.49\linewidth}
  \centering
  \includegraphics[width=1.7in, height=1.5in] {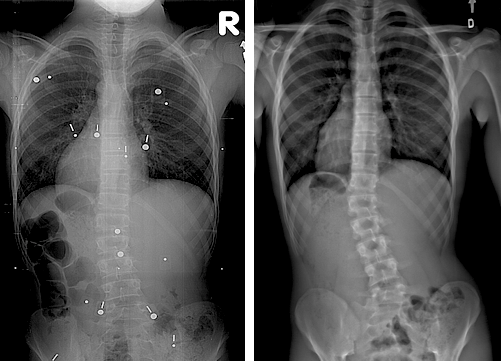}
      \centerline{(a)}
\end{minipage}
\begin{minipage}[b]{0.49\linewidth}
    \includegraphics[width=1.7in, height=1.5in] {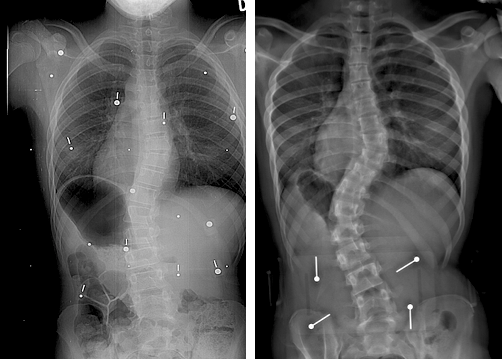}
        \centerline{(b)}
\end{minipage}
   \caption{Examples of similar baseline 3D reconstructions with different longitudinal outcomes. (a) A non-progressive case. (b) A progressive case.}
\label{fig:example}
\end{center}
\end{figure}

Global nonlinear techniques for dimensionality reduction could address these issues \cite{van2009dimensionality} by preserving the global properties of the 3D spine models. Recent studies based on deep learning algorithms such as stacked auto-encoders have successfully represented multiple types of spinal deformations, but was limited to retrospective classification analysis \cite{thong2016three}. On the other hand, prediction models generated from the underlying manifold structure is far from being trivial, and relies on a proper representation of spatiotemporal evolution in an inhomogeneous population with irregular longitudinal evaluation.  Our aim in this paper is to overcome these major limitations by offering a generic framework which captures spatiotemporal variability in a probabilistic embedding. The goal is to predict the curvature evolution from prospectively followed scoliotic patients by using annotated articulated spine models described in manifold space, based on the baseline reconstruction combined with skeletal properties of the patient (Fig. \ref{fig:example}).

Discriminant embeddings take advantage of differences observed between various classes of shapes and create links between disparate data. This factor enables to detect structural alterations at various scales in pathologies \cite{shakeri2015classification}. In fact, typical manifold learning methods tend to better model highly non-linear data, but they do not define distributions over the learned data, providing no measure of confidence in the prediction based on the manifold structure. To address this issue, probabilistic models offer the possibility to establish the relationship between the low-dimensional manifold coordinates and the high-dimensional data, with the assumption that the functions are drawn from Gaussian priors \cite{lawrence2004gaussian}. While previous approaches directly optimized low-dimensional coordinates which limited the evaluation of probabilistic models, recent methods infer locally linear latent variables in order to capture the underlying structure of the manifold \cite{park2015bayesian}.  

Evolution trajectories for growth models have been a popular research topic in neurodevelopment studies for newborns, using geodesic shape regression to compute the diffeomorphisms based on image time series of a population \cite{fletcher2013geodesic, singh2013hierarchical}. These regression models were also used to estimate 4D deformation trajectories by integrating surface information, which would determine the optimal control points and inertia between baseline and longitudinal images through an image-based registration \cite{fishbaugh2014geodesic}. This model showed accurate progression accuracy but required multiple time series, in addition to the baseline image, to perform a prediction. Regression models were proposed for both cortical and subcortical structures, with 4D varifold-based learning framework with local topography shape morphing being proposed by Rekik et al. \cite{rekik2016predicting}, yet there is no framework adapted for progressive spinal curves.

\begin{figure*}[tb]
\begin{center}
\includegraphics[width=5.5in]{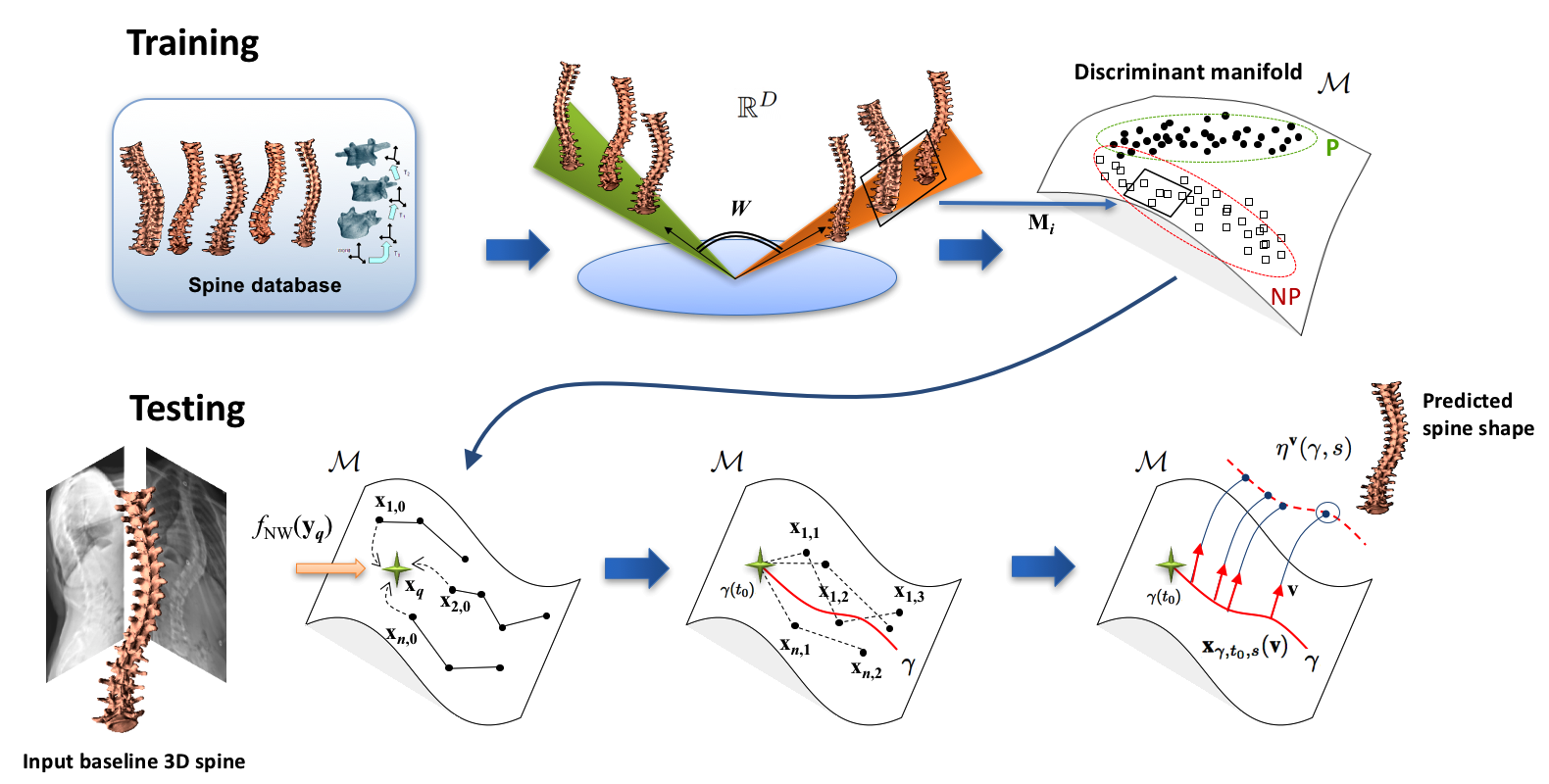}
   \caption{Flowchart diagram of the proposed method. In the training phase, a dataset of spine models are embedded in a discriminant manifold $\mathcal{M}$, distinguishing non-progressive (NP) and progressive (P) samples. During testing, an unseen baseline 3D spine reconstruction $\textbf{y}_q$ is projected on  $\mathcal{M}$ where closest points $\textbf{x}$ are selected to regress the spatiotemporal curve $\gamma$ used for predicting the progression of spinal deformities. }
\label{fig:Flow}
\end{center}
\end{figure*}

This paper presents a prediction framework for the progression of AIS from 3D spine models reconstructed from biplanar X-ray images, which is outlined in Fig. \ref{fig:Flow}. The method first trains a discriminant manifold with Bayesian modeling of input priors using a collection of previously reconstructed 3D spines acquired from longitudinal evaluations of patients with progressive and non-progressive AIS.  A discriminant adjacency matrix is constructed to maximize the separation between these different clinical groups, while minimizing the distance in latent variables belonging to the same class. In the second phase, a new baseline reconstruction is projected onto the manifold, where the neighborhood is identified from the closest samples. A geodesic curve describing spatiotemporal evolution is regressed using discrete approximations, from which the curvature evolution is inferred, yielding a prediction of the intervertebral displacements and shape morphology describing deformation progression.  The method was tested on 133 subjects using personalized 3D reconstructed spine models from biplanar images, with results demonstrating that the discriminatory framework can identify between non-progressive and progressive patients. The contributions of this paper are three-fold: (1) propose a Bayesian manifold learning method which incorporates a discriminant nature to the locally linear latent variable model, exploiting known feature labels from the different classes which are incorporated in the optimization process; (2)  implement a novel discretization procedure of the continuous representation of the geodesic curve, where the approximation is based on samples belonging to the same class; (3) propose a parallel transport curve approach in the tangent space from low-dimensional samples designed for the progression of complex spinal deformity patterns, where a new time-warping function regulating the rate of progression is obtained from flexibility parameters assessed at baseline.


\section{Probabilistic Modeling of Discriminant Spine Manifolds}

The input for training of our predictive framework is a collection of longitudinal articulated spine models which comprises a constellation of vertebral shapes with precise anatomical landmarks located on each vertebra. The same set of landmarks are repeated across all vertebrae. We generate a probabilistic discriminant manifold structure from this training database to differentiate progressive and non-progressive curves by mapping the training set into a simplified manifold domain the dimensionality of which represents to the size of admissible variations. 

\subsection{Spine Model Description}
\label{sec.articulatedrep}
The spine model is described by $\textbf{S}= \{\mathbf{s}_{1},\ldots,\mathbf{s}_{L}\}$, which includes an articulation of $L$ vertebrae. For each vertebra $\mathbf{s}_{i}$, a geometric representation of the $i$th vertebra is obtained by generating a surface model where surface positions are corresponding from one shape to another. Furthermore, each personalized model $\mathbf{s}_{i}$ includes a set of anatomical points which are used to perform a point-based transform of the vertebra onto it's superior level. In order to account for morphological differences between vertebral levels, the Procrustes alignment superimposition for vertebral shapes \cite{korez2015framework} is used to determine the transformation for all inter-vertebral levels. This superimposition is used to establish the registration matrix and determine the orientation and translation parameters. Therefore, the global spine shape model is represented as a vector of inter-vertebral registrations assigned to each vertebral level $ [T_{1}, T_{2}, \ldots , T_{L}]$ between each vertebra. To perform global shape modeling of the shape $\textbf{S}$, we use an absolute representation by representing each transformation as a combination of previous transforms:
\begin{equation}\label{eq.Aabs}
\textbf{y}_{\textrm{i}} = [T_{1}, \mathbf{s}_{1}; T_{1} \circ T_{2}, \mathbf{s}_{2},\ldots , T_{1} \circ T_{2} \circ \ldots \circ T_{L},\mathbf{s}_{L}]
\end{equation}
using recursive compositions. The feature vector $\textbf{y}_{\textrm{i}}$ controls the position and orientation of the object constellation, while describing the shape model $\textbf{S}$ capturing vertebral morphology. The model can is deformed by applying displaced to the inter-vertebral parameters. By extending this to the entire absolute vector representing the spine model, this then achieves a global deformation. In this case, registrations are described in the reference coordinate system of the lower vertebra, corresponding to it's principal axes of the cuneiform shape with the origin positioned at the center of mass of the vertebra. The rigid transformations are the combination of a rotation matrix $R$ and a translation vector $t$. We formulate the rigid transformation $T = \{R, t\}$ of a vertebral model $\mathbf{s}_{i}$ as $y = Rx+t$ where $x, y, t  \in \Re^{3}$. Composition is given by $T_{1}\circ T_{2}=\{R_{1}R_{2},R_{1}t_{2}+t_{1}\}$.

\subsection{Probabilistic Model for Discriminant Manifolds}

Manifold learning algorithms are based on the premise that data are often of artificially high dimension $D$ and can be embedded in a lower dimensional $d$ space. On the other hand, the presence of multiple classes and data points falling outside the normal distribution can alter the discriminative behavior of the model. We propose to learn the optimal separation between two classes (1) non-progressive (NP) AIS patients and (2) progressive (P) AIS patients, by using a discriminant graph-embedding. Here, $n$ labelled points  $\mathbb{Y}=\{(\textbf{y}_i,l_i,t_i)\}_{i=1}^{n}$ defined in  $\mathbb{R}^{D}$ are generated from the underlying manifold $\mathcal{M}$, where $l_i$ denotes the label (NP or P) and $t_i$ represents the time of observation. For the labelled data, there exists a low-dimensional (latent) representation of the high-dimensional samples such that  $\mathbb{X}=\{(\textbf{x}_i,l_i,t_i)\}_{i=1}^{n}$ defined in $\mathbb{R}^{d}$. We assume here that the mapping $\textbf{M}_i \in \mathbb{R}^{D \times d}$ between high and low-dimensional spaces is locally linear, such that tangent spaces in local neighbourhoods can be estimated with ${ \textbf{y}_j - \textbf{y}_i}$ and ${ \textbf{x}_j - \textbf{x}_i}$, representing the pairwise differences between connected neighbours ${i,j}$. Therefore the relationship can be established as $ \textbf{y}_j - \textbf{y}_i \approx \textbf{M}_i ( \textbf{x}_j - \textbf{x}_i )$.

In order to generate the embedding in $\mathbb{R}^{d}$,  the local structure of the data needs to be maintained in the new embedding. The graph $\mathcal{G}=(\textbf{\emph{V}},\textbf{\emph{W}})$ is an undirected similarity graph, with a collection of nodes $\textbf{\emph{V}}$ connected by edges, and the symmetric matrix $\textbf{\emph{W}}$ describing the edges between the nodes of the graph. The diagonal matrix $\textbf{\emph{D}}$ and the Laplacian matrix $\textbf{\emph{L}}$ are defined as $\textbf{\emph{L}}= \textbf{\emph{D}} - \textbf{\emph{W}}$, with $\textbf{\emph{D}}(i,i) = \sum_{j \neq i}\textbf{\emph{W}}_{ij} \forall i$.

Using the framework from Park et al.  \cite{park2015bayesian}, we can determine a distribution of linear maps associated with the low-dimensional representation to describe the data likelihood:

\begin{equation}\label{eq.costlle}
\log p( \mathbb{Y} | \mathcal{G}) = \log \iint   p ( \mathbb{Y}, \textbf{M},  \mathbb{X}| \mathcal{G}) d\textbf{x}d\textbf{M}.
\end{equation}

This joint distribution can be separated into three prior terms: the linear maps, latent variables and the likelihood of the high dimensional points $ \mathbb{Y}$:

\begin{equation}\label{eq.costlle}
 p(  \mathbb{Y}, \textbf{M},  \mathbb{X} | \mathcal{G}) =   p (  \mathbb{Y} | \textbf{M},  \mathbb{X}, \mathcal{G}) p(\textbf{M} | \mathcal{G}) p( \mathbb{X}| \mathcal{G}).
\end{equation}
We now define the neighborhood selection used to establish the discriminant similarity graphs, as well as define each of the three prior terms included in the joint distribution.
\\
\\
\subsubsection{Nearest neighbor selection}
An important drawback of embedding algorithms is the underlying assumption that the similarity between embedded samples can be estimated by the use of Euclidean distances. In this paper, a similarity measure based on the domain of articulated structures is used to handle the anatomical spine variability in the pathological population \cite{Kadoury10Miccai}. It is anchored on the natural properties of Riemannian manifold geometry which enables discriminating between inter-vertebral vectors independently of the overall manifold structure. For each sample, the closest neighbor can be found with a geodesic metric, and is defined as $d_{\mathbb{M}}(\textbf{y}_{\textrm{i}},\textbf{y}_{\textrm{j}})$ which finds the deviation of spine vector articulations $i,j$, with $\textbf{y}_{\textrm{i}}$ and $\textbf{y}_{\textrm{j}}$ as the feature vectors described in Eq.(\ref{eq.Aabs}). The deviation metric is then described as the summation of deviations in transformations:
\begin{align}\label{eq.4}
d_{\mathbb{M}}(\textbf{y}_{\textrm{i}},\textbf{y}_{\textrm{j}}) &= \sum_{k=1}^{L} d_{\mathbb{M}}(T_{k}^{i},T_{k}^{j}) \\
&=\sum_{k=1}^{L} d_{\mathbb{M}}
\begin{pmatrix}
\begin{pmatrix}
R_{k}^{i} & t_{k}^{i}\\
0 & 1
\end{pmatrix}
,
\begin{pmatrix}
R_{k}^{j} & t_{k}^{j}\\
0 & 1
\end{pmatrix}
\end{pmatrix} \nonumber
\end{align}
where the canonical representation encodes the intrinsic ($t$) and orientation ($R$) parameters. The difference between analogous articulations is computed within the geodesic framework such that:
\begin{equation}\label{eq.dismetric}
d_{\mathbb{M}}(\textbf{y}_{\textrm{i}},\textbf{y}_{\textrm{j}})=\sum_{k=1}^{L}\|t_{k}^{i}-t_{k}^{j}\|+\sum_{k=1}^{L}d_{G}(R_{k}^{i},R_{k}^{j}).
\end{equation}
The first term evaluates intrinsic distances in the $L_{2}$ norm. Using the geodesics, it is possible to define a diffeomorphism between rotation neighborhoods in $\mathcal{M}$ and a tangent plane $T_x\mathcal{M}$. The exponential map at $x \in \mathcal{M}$ maps vectors of the tangent plane $T_x\mathcal{M}$ to a point in the manifold which is reached by the geodesic $\gamma_{x,v}$ in a unit time. In other words, if $\gamma_{(x,v)}(1) = y$, then the inverse mapping is known as $\log_{x}(y) = v$. The distances are therefore computed with the following norm $d_{G}(R_{k}^{i},R_{k}^{j})=\|\log((R_{k}^{i})^{-1}R_{k}^{j})\|_{F}$ based on the geodesic distances ($d_{G}$) in the 3D manifold. This is feasible since rotations $R_{k}^{i},R_{k}^{j}$ are nonsingular, invertible matrices. Articulation differences between analogous vertebrae are computed instead of vertebra shape variations since the goal of this step is to capture the differences in pose between the different spine samples in the dataset. In a previous study \cite{kadoury2013spine}, we demonstrated that using an articulated deviation metric was sufficiently accurate to capture the differences in manifold space. We can now proceed to the manifold reconstruction using the local support in high-dimension data.

\subsubsection{In-class and between-class graphs}
The geometrical structure of the manifold $\mathcal{M}$ is determined by constructing a within-class similarity graph $\textbf{\emph{W}}_w$ for feature vectors in the same class and a between-class similarity graph $\textbf{\emph{W}}_b$, to separate features from both classes. At the time of building the discriminant locally linear latent variable embedding, elements are partitioned into $\textbf{\emph{W}}_w$ and $\textbf{\emph{W}}_b$. As a first step, the graph model $\mathcal{G}$ is determined by linking edges only to points belonging to a particular group (e.g.  NP). Second, individual points are reconstructed based on feature vectors included in the same class. Local coefficients used for the reconstruction of samples integrated in the graph $\textbf{\emph{W}}_w$ are defined as:
\begin{equation}\label{eq.simwithin}
\textbf{\emph{W}}_{w_{i,j}}=
\begin{cases}
1  \,\,  &\text{if} \,\,\ \textbf{y}_i \in \mathcal{N}_w(\textbf{y}_j) \,\, \text{or} \,\,  \textbf{y}_j \in \mathcal{N}_w(\textbf{y}_i)\\
0, \,\, &\text{otherwise.}
\end{cases}
\end{equation}	
with $\mathcal{N}_w$ containing neighbors of the same class. Conversely, $\textbf{\emph{W}}_b$ represents the edge properties which are highly penalized during the inference step.  Local coefficients used for mapping samples further away are obtained with:
\begin{equation}\label{eq.simbetween}
\textbf{\emph{W}}_{b_{i,j}}=
\begin{cases}
1  \,\,  &\text{if} \,\, \textbf{y}_i \in \mathcal{N}_b(\textbf{y}_j) \,\, \text{or} \,\,  \textbf{y}_j \in \mathcal{N}_b(\textbf{y}_i)\\
0, \,\, &\text{otherwise}
\end{cases}
\end{equation}	
with $\mathcal{N}_b$ containing samples having different class labels from the $i$th sample. Both $\mathcal{N}_b$ and $\mathcal{N}_w$ neighborhoods are determined from the closest samples as determined by the metric in  Eq.(\ref{eq.dismetric}). The objective is to transform points to a new manifold $\mathcal{M}$ of dimensionality $d$, i.e. $\textbf{y}_i \rightarrow \textbf{x}_i$, by mapping connected samples from the same group in $\textbf{\emph{W}}_w$ as close as possible to the class cluster, while moving NP and P samples of $\textbf{\emph{W}}_b$ as far away from one another as possible. 
\\
\\
\subsubsection{Model components}
The prior added on the latent variables $\mathbb{X}$ are located at the origin of the low-dimensional domain, while minimizing the Euclidean distance of neighboring points that are associated with the neighborhood of high-dimensional points and maximizing the distance between coordinates of different classes. In order to set the variables with an expected scale $\sigma$ and  $H$ representing the probability density function, the following log prior is defined:
\begin{align}\label{eq.costlle}
\log p(\mathbb{X} | \textbf{\emph{W}}, \sigma ) =  -\frac{1}{2} &\sum_{i=1}^{n} ( \sigma \|  \textbf{x}_i \| + \sum_{j=1}^{n} \textbf{\emph{W}}_{w_{i,j}} \| \textbf{y}_i - \textbf{y}_j \|^2  -  \nonumber  \\ 
&\sum_{j=1}^{n} \textbf{\emph{W}}_{b_{i,j}} \| \textbf{y}_i - \textbf{y}_j \|^2 ) - \log H_{\mathbb{X}}.  
\end{align}

The prior added to the linear maps defines how the tangent planes described in low and high dimensional spaces are similar based on the Frobenius norm $F$. This prior ensures smooth manifolds:

\begin{align}\label{eq.costlle}
\log p(  \textbf{M}  | \textbf{\emph{W}} ) &=  -\frac{1}{2} \Bigg (  \bigg \| \sum_{i=1}^{n}   \textbf{x}_i \bigg\| ^2_{F} - \\   
&\sum_{i=1}^{n} \sum_{j=1}^{n}  \big ( \textbf{\emph{W}}_{w_{i,j}}  - \textbf{\emph{W}}_{b_{i,j}} )  \|\textbf{M}_i - \textbf{M}_j \|^2_{F}   \Bigg ) -  \log H_{ \textbf{M}}. \nonumber
\end{align}

Finally, approximation errors from the linear mapping $\textbf{M}_i$ between low and high-dimensional domains are penalized by including the following log likelihood:

\begin{align}\label{eq.costlle}
\log p(\mathbb{Y}| \mathbb{X},  \textbf{W}, \omega) &= \| \sum_{i=1}^{n}  \textbf{y}_i \|^{2} \\
&- \frac{1}{2}  \sum_{i=1}^{n} \sum_{j=1}^{n} \textbf{\emph{W}}_{w_{i,j}} \Delta(i,j)^{\text{T}}\omega_w \textbf{I} \Delta(i,j) \nonumber \\
 & + \frac{1}{2}  \sum_{i=1}^{n} \sum_{j=1}^{n} \textbf{\emph{W}}_{b_{i,j}} \Delta(i,j)^{\text{T}}\omega_b \textbf{I} \Delta(i,j) -\log H_{ \textbf{y}}\nonumber
\end{align}

with $\Delta(i,j)$  the difference in Euclidean distance between pairs of neighbors in high and low-dimensional space and $\omega$ the update parameters for the EM inference. Samples of $\textbf{y}$ are drawn from a multivariate normal distribution. 

\subsection{Variational Inference}

The objective of the last step is to infer the low-dimensional coordinates and linear mapping function for the described model, as well as the intrinsic parameters of the model $\Phi=(\sigma,\omega)$. This is achieved by maximizing the marginal likelihood of:

\begin{equation}\label{eq.marglikelihood}
\log p( \mathbb{Y} | \textbf{W}, \Phi )  = \iint  \rho (\textbf{M}, \mathbb{X} ) \log \frac{p ( \mathbb{Y}, \textbf{M} , \mathbb{X} | \textbf{W}, \Phi)}{\rho (\textbf{M}, \mathbb{X} ) } d\textbf{x}d\textbf{M}.
\end{equation}

By assuming the posterior  $\rho (\textbf{M}, \mathbb{X} )$ can be factored in separate terms  $\rho (\textbf{M})$ and $\rho( \mathbb{X} )$, a variational expectation maximization algorithm can be used to determine the model's parameters, which are initialized with $\Phi$. The E-step updates the independent posteriors $\rho(\mathbb{X})$ and $\rho(\textbf{M})$, while the parameters of $\Phi$ are updated in the M-step by maximizing Eq. (\ref{eq.marglikelihood}).

The discriminant latent variable model can then be used to obtain the low-dimensional representation of a feature vector. The variational EM algorithm described in the previous section can be used to transform a set of new input points $\textbf{y}_q$ without changing the overall neighborhood graph structure, by finding the distribution of the local linear map $\textbf{y}_q$ and it's low-dimensional coordinate using the E-step explained above. Once the manifold representation $\textbf{x}_q$ is obtained, a cluster analysis finds the corresponding class in the manifold, yielding a prediction of the input feature vector $\textbf{y}_q$.

\section{Progression Model for Spinal Deformities}
Once the appropriate shape variations are determined from the probabilistic modeling of spine progression in a discriminant embedding, a new 3D spine can be classified in P and NP classes, and subsequently predict it's progression. During testing, a new baseline reconstruction is given and prediction of progression is obtained by first projecting this baseline 3D reconstruction onto the manifold to identify the neighborhood from the closest samples (Sec. III-A). A geodesic curve describing spatiotemporal evolution is then regressed using discrete approximation to infer the curvature evolution for a prediction of progressive spinal deformities (III-B). The prediction of a spine at a given point in time is obtained by performing the inverse transformation, using the exponential mapping function, from a given point on the regressed curve, to the high-dimensional space (III-C).

\subsection{Baseline Shape Projection on $\mathcal{M}$}
To obtain the embedded point from a new 3D spine model described in radiographic space, the low-dimensional representation needs to be determined based on its intrinsic coordinates. By assuming there exists a forward mapping $f : \mathbb{R}^{D} \rightarrow  \mathcal{M}$ linking real-world in $\mathbb{R}^{D}$ to the sub-space $\mathcal{M}$, which can be obtained from the joint distribution of the $\mathbb{R}^{D}$ and $\mathcal{M}$ relationship, we can create a continuous and regular kernel that is defined in the local vicinity of a query point. By following the conditional expectation theory, the manifold-based function is defined as:
\begin{equation} \label{eq.conditionalexp}
f(\textbf{y}_{i})\equiv E( \textbf{x}_{i} |\mathcal{M}^{-1}(\textbf{x}_{i})=\textbf{y}_{i})=\int \textbf{x}_{i} \frac{p(\textbf{y}_{i},\textbf{x}_{i})}{p_{\mathcal{M}(\textbf{x}_{i})}(\textbf{y}_{i})}d\text{d}
\end{equation}
which describes the regional variation of samples in $\mathbb{R}^d$. Here, both $p_{\mathcal{M}(\textbf{x}_{i})}(\textbf{y}_{i})$ (marginal density of $\mathcal{M}(\textbf{x}_{i})$) and $p(\textbf{y}_{i},\textbf{x}_{i})$ (joint density) need to be found. Following the \emph{Nadaraya-Watson} kernel regression  \cite{Nadaraya64}, the marginal and joint terms can be estimated with  kernel functions such that  $p_{\mathcal{M}(\textbf{x}_{i})}(\textbf{y}_{i})=\frac{1}{K}\sum_{j\in \mathcal{N}(i)}G_h(\textbf{y}_{i},\textbf{y}_{j})$ and $p(\textbf{y}_{i},\textbf{x}_{i})=\frac{1}{K}\sum_{j\in \mathcal{N}(i)}G_h(\textbf{y}_{i},\textbf{y}_{j})G_g(\textbf{x}_{i},\textbf{x}_{j})$ following a conditional expectation setting \cite{Davis07}. The Gaussian regression kernels $G$ require the neighbors $ \textbf{x}_{j}$ of $j \in \mathcal{N}(i)$ to determine the bandwidths $h,g$ so it includes all $K$ data points ($\mathcal{N}(i)$ representing the neighborhood of $i$) selected only from progressive samples at baseline ($t=0$), as the progression prediction is not performed for non-progressive cases. Plugging these estimates in Eq.(\ref{eq.conditionalexp}), this gives:
\begin{equation}\label{eq.firstnadaraya}
f_{\textmd{NW}}(\textbf{y}_{i})=\int \textbf{x}_{i} \frac{\tfrac{1}{K} \sum_{j\in \mathcal{N}(i)} G_h(\textbf{y}_{i},\textbf{y}_{j})G_g(\textbf{x}_{i},\textbf{x}_{j})} {\tfrac{1}{K}  \sum_{j \in \mathcal{N}(i)} G_h(\textbf{y}_{i},\textbf{y}_{j})}d\text{d}.
\end{equation}
By assuming $G$ is symmetric about the origin, we propose to integrate in the kernel regression estimator, the geodesic distance on the manifold $d_{\mathbb{G}}$ which is particularly suited for articulated diffeomorphisms.  This generalizes the expectation such that the observations $\textbf{y}$ are defined in manifold space $\mathcal{M}$:
\begin{equation}\label{eq.finalnadayara}
f_{\textmd{NW}}(\textbf{y}_{i})=\underset{\textbf{x}_{i}}{\operatorname{argmin}} \frac{\sum_{j\in \mathcal{N}(i)}G(\textbf{x}_{i},\textbf{y}_{j})d_{\mathbb{G}}(\textbf{x}_{i},\textbf{y}_{j})}{\sum_{j \in \mathcal{N}(i)} G(\textbf{y}_{i},\textbf{y}_{j})}
\end{equation}
which integrates the geodesic distance metric $d_{\mathbb{G}}(\textbf{x}_{i},\textbf{y}_{j})$ which is defined in manifold space and updates $f_{\textmd{NW}}(\textbf{y}_{i})$ from neighboring points of $\textbf{x}_i$ found from ambient domain. The kernel is therefore restrained for samples points which exhibit the same morphology as the vicinity around $\textbf{y}_{i}$ is the same as for $\textbf{x}_{i}$.

\subsection{Spatiotemporal Model from Manifold Space}

Given a query baseline sample $\textbf{y}_q$ with it's projection $\textbf{x}_q$ from Eq. (\ref{eq.finalnadayara}), we are looking for a geodesic curve $\gamma : [t_1,t_N]$ which represents the spatiotemporal evolution for a given time interval that adequately represents the fitted data, while maintaining a regular and smooth shape along the manifold space. The embedded data provided from the $K$ individuals identified earlier with progressive deformations belonging to $\mathcal{N}(\textbf{x}_q)$, measured at different time points, creates a low-dimensional Riemannian manifold where data points $\textbf{x}_{i,j}$, with $i$ denoting a particular individual and $j$ the time-point measurement. Here, $j=0$ represents the baseline reconstruction of the patient. We assume here that the discriminant manifold is complete with geodesics defined for all time.  We also assume that the geodesic curve can be defined as a regression problem for time-labeled data in $\mathbb{R}^{D}$ in the continuous setting, such that a smooth curve can be defined over the time interval with points defined in $\mathbb{R}^{d}$.

The main challenge in the continuous representation of the curve lies in the fact that the problem is a variational problem in infinite dimension.
A discretization scheme for implementation purposes is therefore necessary for an appropriate application. We use the discretization procedure proposed by Boumal et al. \cite{boumal2011discrete}, such that:
\begin{align}\label{eq.discreteminimization}
E(\gamma) = \dfrac{1}{2} \sum_{i \in \mathcal{N}(\textbf{x}_q)} &\sum_{j=0}^{t_N} w_i \| \gamma(t_{i,j}) - (\textbf{x}_{i,j} - (\textbf{x}_{i,0} - \textbf{x}_q)) \|^2 \nonumber \\
&+ \dfrac{\lambda}{2}  \sum_{i=1}^{K_d} \alpha_i \|v_i\|^2 + \dfrac{\mu}{2}  \sum_{i=1}^{K_d} \beta_i \| a_i \|^2 
\end{align}
which reduces the problem to a highly structured quadratic optimization problem without constraints, and can be solved using  LU decomposition and substitutions of the singular terms, with $K_d$ the number of discretized points along $\gamma$.
The first term is a misfit penalty which measures the geodesic distance on $\mathcal{M}$ between true embedded coordinates $\textbf{x}_{i,j}$ and attempts to obtain the best fit between the regressed curve and actual data points, weighted by variables $w_i$ based on the distance between samples. That means that the fitted curve will lie as close as possible to the points $\textbf{x}_{i,j}$, which are shifted by $\textbf{x}_q$ so that baseline samples are co-aligned. The second term is the velocity penalty, which seeks to minimize the $L_2$ norm of the first derivative of the regressed curve $\gamma$. This term seeks to have the derivatives of the curve with a lower norm value, thus avoiding hard transitions or highly curved sections, and is regulated by $\alpha_i$. 
The third term is the acceleration penalty term, minimizing the $L_2$ norm of the second derivative of the regressed curve $\gamma$, and is regulated by $\beta_i$. The tangent vectors $v_i$ and $a_i$, rooted at $\gamma_i$, are approximations to the velocity and acceleration vectors $\dot{\gamma}(t_i)$ and $\dfrac{\text{D}^2\gamma}{\text{d}t^2} (t_i)$, respectively. The estimates for both velocity and acceleration, weighted by parameters $\{ \lambda,\mu \}$ are obtained from geometric finite differences which determine the backward and forward step-sizes along $\gamma$, which could be interpreted as direction vectors in manifold space. For more detail, reader should refer to Boumal et al. \cite{boumal2011discrete}.

In order to avoid convergence problems and slow optimization using steepest descent algorithms, we resort to a second-order method to minimize $E(\gamma)$. We use a non-linear conjugate gradient method defined in the low-dimensional space $\mathbb{R}^{d}$. We therefore define $\gamma$ as the curve defined in $\mathcal{M}$ for all time, with a time point $t_0$. The curve creates a group average of spatiotemporal transformations based on individual progression trajectories. 

\subsection{Morphology Prediction of Spine Deformation}

In the last step of the testing phase, we use the spatiotemporal evolution using the geodesic curve $\gamma : \mathbb{R}^{D}  \rightarrow \mathcal{M}$ determined previously, where for each point $\textbf{x} \in \mathcal{M} $ on the manifold, it has a vector $\textbf{v}$ associated with it in the tangent plane, such that $\textbf{v} \in \text{T}_{\textbf{x}} \mathcal{M}$. Using Riemannian exponential theory, a mapping can be estimated such that $ \text{Exp}_{\textbf{x}}^{\mathcal{M} }(\textbf{v})$, i.e. the point at $t=1$ from the geodesic starting at $\textbf{x}$ with velocity $\textbf{v}$. Based on this mapping function, we use the concept of parallel transport curves in the tangent space from low-dimensional manifolds proposed by  Schiratti et al. \cite{schiratti2015learning}, which maps a series time-index vectors on the tangent spaces along $\gamma$, thus creating parallel curves which are described in the ambient space as shown in Fig. \ref{fig:Flow}, modeling shape changes in $\mathbb{R}^{D}$. The key idea is that by navigating the spatiotemporal geodesic curve modeling progression in time, we can derive the appearance at various time-points in $ \mathbb{R}^{D}$ by exponential mapping. In order to ensure the parallel transport defines a spatiotemporal evolution in the coordinate system of the spine baseline, the tubular neighborhood theorem is used based on ICA \cite{hyvarinen2004}.
Given the regressed spatiotemporal curve $\gamma$, the manifold at time point $t_0$ with a vector $\textbf{v}$ associated with the tangent plane at $\gamma(t_0)$, we can therefore define the parallel curve:
\begin{equation}\label{eq.parallelcurve}
\eta^{\textbf{v}} (\gamma,s)= \text{Exp}_{ \gamma(s)}^{\mathcal{M}} (\textbf{x}_{\gamma,t_0, s}(\textbf{v})), \,\,\, s \in \mathbb{R}^{d}.
\end{equation}
By generalizing this concept and repeating the mapping along $\gamma(s)$, we can create a model built from the manifold points seen as samples of individual progression trajectories. This maps points along $\gamma(s)$ to create new points $\eta^{\textbf{v}} (\gamma,\cdot)$ which are parallel to the embedded geodesic curve in $\mathcal{M}$, thus describing the spatiotemporal variation in ambient space.

We can now define a time warp function $\phi_i(t)= C_i (t-t_0-\tau_i) +t_0$ which allows to vary $s$ along the spatiotemporal curve. The time-warp function includes a patient specific acceleration factor which encodes the flexibility of the spine based on spine bending radiographs. This was calculated by the ratio $C_i$ of the Cobb angle difference between standing and bending films, with the initial Cobb angle \cite{lamarre2009}. This encodes whether the patient is progressing faster or slower than the group of $K$ samples. The time-shift parameter $\tau_i$ enables to encode the relative difference of the particular sample $i$ with respect to the group average of the regressed curve.

For spine progression estimation, space-shift vectors $\textbf{v}_i$ are determined by the principal direction of the hyperplane perpendicular to the tangent plane $\text{T}_{\textbf{x}_i} \mathcal{M}$ in low-dimensional space via an eigendecomposition \cite{schiratti2015learning}.
Therefore for a new mapped point $\textbf{x}_q$ which represents the embedded representation of the baseline 3D reconstruction, and a future time point $t_k$ with  the regressed geodesic curve obtained from the manifold points $\textbf{x}$ in $\mathcal{N}(\textbf{x}_q)$, the predicted models can be described as:
\begin{equation}\label{eq.spatiotemporalmodel}
\textbf{y}_{q,t_k} = \eta^{\textbf{v}_q} (\gamma,\phi_i(t_{k})) + \epsilon_{q,t_k}
 \end{equation}
with $\epsilon_{q,t_k}$ a zero-mean Gaussian distribution. This yields an output $\textbf{y}_{q,t_k}$  which is the predicted shape model that is generated using the proposed model, which is described in the ambient space $\mathbb{R}^{D}$. This output describes the articulated pose estimation at a time-point $t_k$, along with the shape model $\textbf{S}$, i.e. a constellation of inter-connected vertebra models, each annotated with characteristic anatomical landmarks, describing local shape variations caused by the progression of the spine deformation.

\section{Results}
\label{Results}

\subsection{Training Data}

The discriminant manifold was built from a database containing $745$ 3D spine reconstructions, originating from $133$ patients demonstrating several types of deformities. Patients were recruited from a single center prospective study, with the inclusion criteria being evaluated by an orthopedic surgeon and a main curvature angle between 11$^{\circ}$ and 40$^{\circ}$. Patients were divided in two groups based on the severity of the main curve, with the first group composed of 52 progressive patients with a difference of over 6$^{\circ}$ between the first and last visits. The second group was composed of 81 nonprogressive (NP) patients with a difference of 6$^{\circ}$ or less between baseline and longitudinal scans (up to 3 years after baseline). This threshold was selected based on the level of confidence for radiographic measurements. 

The database is composed of  3D spine reconstructions obtained from biplanar radiographic images \cite{Kadoury07mbec}. Each model includes 12 thoracic and 5 lumbar vertebrae, each vertebra annotated with 4 pedicle tips and 2 center points placed on the vertebral endplates, and validated by an experienced radiologist. Using these precise landmarks as reference points, triangulated vertebra shape models were generated using templates obtained from CT images of a cadaveric spine model. CT images were acquired with 1mm slice spacing. Deformation of the template to the patient-specific landmarks was obtained using a B-spline FFD. The templates included all 17 vertebrae, with an average $5226$ vertices on each model. On each template, the exact 6 anatomical landmarks described previously were located. These landmarks were also used to establish the local coordinate system of the vertebra, describing the orientation and location (ground truth pose).


\subsection{Manifold Generation}
\begin{figure}
\begin{center}
\includegraphics[width=0.7\linewidth]{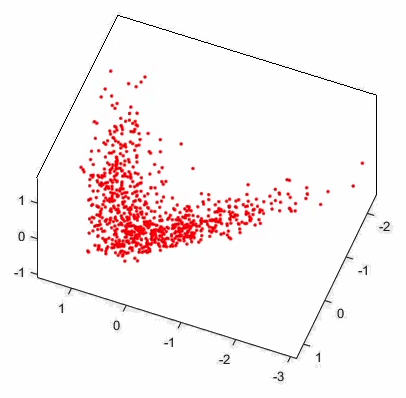}
\end{center}
   \caption{Discriminant manifold embedding of the spine dataset comprising 745 models exhibiting various types of deformities. The sub-domain was used to  distinguish progressive scoliotic patients from non-progressive patients, and uses the low-dimensional feature coordinates to estimate a spatiotemporal evolution model.}
\label{fig:Finalmanifold}
\end{figure}

As described in the manifold description, the parameter $d$ which dictates the extent of the low-dimensional sub-space has a strong influence on the precision of the predicted spine models obtained from longitudinal data. We found that the trend of the nonlinear residual reconstruction error stabilized for the entire training set at $d=8$. The optimal neighborhood size ($K=10$) was found based on the significant edges in the similarity graphs  $\textbf{\emph{W}}$. Because the P group was much more sparse with samples being increasingly spread out than the NP group that tended to be more uniform, the optimal compromise was found when $\omega_w=0.3$  and $\omega_b=0.7$. Fig. \ref{fig:Finalmanifold} displays the  embedding from the training data of spine models in $\mathcal{M}$. 
\\
We then tested the forward mapping function, projecting a baseline 3D model of the spine onto the discriminant manifold. For each sample in the training dataset using a leave-one-out procedure, we obtained the forward transformation and evaluated the prediction along the geodesic transport curves when $t=0$, by comparing the model to the ground-truth reconstruction. In Table \ref{table:identend}, we present the quantitative evaluation using three error metrics, namely the angular error (AE) measured in degrees, the magnitude of differences (MOD), similar to maximal differences measured in millimeters and mean centroid distance (MCD). Results obtained from two other kernel functions (Fisher and Radial Basis Functions) in addition to the Riemannian kernel function (RKF) described in this work were used in order to asses the performance of the regression based in a conditional expectation setting. 

Results were categorized in 5 different classes corresponding to different types of deformity. All error measures were lower with the RKF kernel, with the MOD metric significantly ($p < 0.05$) lower to the two other kernels. The method performs particularly well for severe deformations such as C4. This confirms the advantage of integrating geodesic distance metrics between sample points based on articulation distortions when estimating joint and marginal densities.

\begin{table}
\renewcommand{\arraystretch}{1.3}
\caption{Differences in articulation vectors for each case in the training dataset by comparing projection kernels, grouped into 5 classes of deformation models. Patients were classified as normal (C1), right-thoracic (C2), left-lumbar (C3), right-thoracic-left-lumbar (C4) and left-thoracic (C5).}
\hfill{}
\scalebox{0.85}{
\begin{tabular}{l||c|c|c|c|c|c|c|c|c}
\hline
\textbf{}& \multicolumn{3}{c|}{MOD ($T_{t}$) (mm)  } & \multicolumn{3}{c|}{AE ($T_{R}$) (deg) } & \multicolumn{3}{c}{3D MCD (mm)}\\
\cline{2-10}
\textbf{}& RBF & Fisher &  \textbf{RKF} & RBF & Fisher &  \textbf{RKF} & RBF & Fisher &  \textbf{RKF} \\
\hline
\textbf{C1}  & 0.78 & 0.80 & \textbf{0.38} & 0.48 & 0.49 & \textbf{0.35} & 0.46 & 0.47 & \textbf{0.36}\\
\hline
\textbf{C2}    & 1.84 & 2.30 & \textbf{0.73} & 1.10 & 1.54 & \textbf{0.43} & 1.05 & 1.45 & \textbf{0.69}\\
\hline
\textbf{C3}                  & 1.05 & 1.01 & \textbf{0.60} & 0.92 & 0.98 & \textbf{0.61} & 0.85 & 0.77 & \textbf{0.58}\\
\hline
\textbf{C4}                  & 2.12 & 2.47 & \textbf{0.84} &  0.95 & 1.28 & \textbf{0.61} & 1.16 & 1.50 & \textbf{0.77}\\
\hline
\textbf{C5}                  & 2.33 & 1.44 & \textbf{0.41} & 1.23 & 1.51 & \textbf{0.95} & 0.94 & 1.02 & \textbf{0.55}\\
\hline
\end{tabular}}
\hfill{}
\label{table:identend}
\end{table}

\subsection{Classification Framework}

\begin{figure*}[tb]
\begin{minipage}[b]{0.3\linewidth}
  \centering
  \includegraphics[height=1.9in] {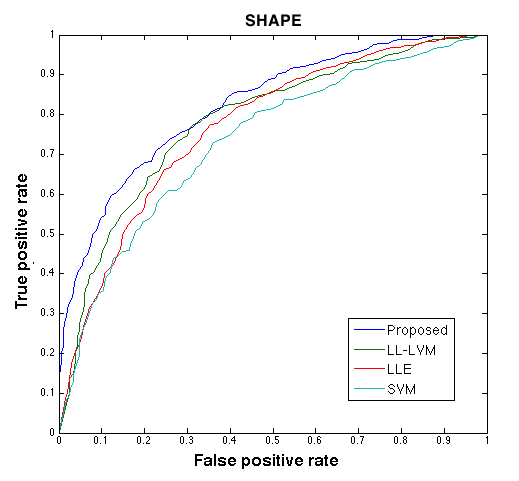}
\end{minipage}
\begin{minipage}[b]{0.3\linewidth}
  \centering
  \includegraphics[height=1.9in] {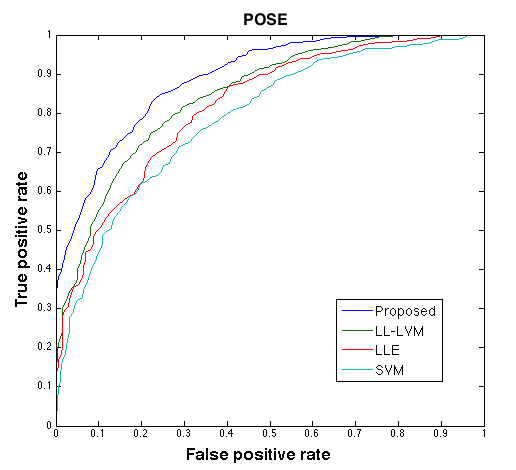}
\end{minipage}
\begin{minipage}[b]{0.3\linewidth}
  \centering
  \includegraphics[height=1.9in] {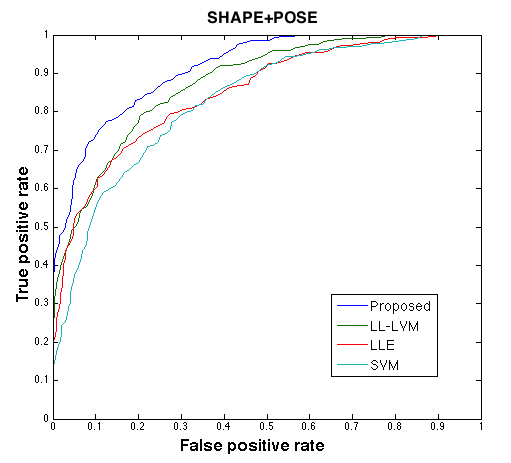}
\end{minipage}

\caption{ROC curves comparing the SVM, LLE and LL-LVM with the proposed method for NP/P prediction using only shape, only inter-vertebral (IV) poses and combining both shape and IV poses.}
\label{fig:ROCcurves}
\end{figure*}

We then tested the classification framework as presented in Section II. Here, a 9-fold cross-validation was performed to assess the performance of the method. This means that the cohort  was divided into 9 equal sets, partitioned randomly. At each run, training was performed on 8 folds (188 cases in total), and validated on the remaining fold. This strategy was repeated 9 times. We evaluated the classification accuracy for discriminating between NP and P scoliotic patients using the baseline 3D reconstructions, by training the model using only vertebral shapes, only inter-vertebral (IV) poses and with a combination of both shape+IV poses. Fig. \ref{fig:ROCcurves} presents ROC curves obtained by the proposed and comparative methods such as SVM (nonlinear RBF kernel), locally linear embedding (LLE) and  locally linear latent variable model (LL-LVM) \cite{park2015bayesian}. The discriminative nature of the proposed framework clearly shows an improvement to standard learning approaches models which were trained using shape only, IV poses only and combined  features. Table \ref{table2} presents accuracy, sensitivity and specificity results for classification between NP and P patients.  It illustrates that increased accuracy (81.0\%) can be achieved by combining shape and IV pose features, showing the benefit of extracting complementary features from the dataset for prediction purposes. We did not observe significant differences in accuracy using a leave-one-out strategy. When comparing the performance of the proposed method to the other learning methods (SVM, LLE, LL-LVM), the probabilistic model integrating similarity graphs shows a statistically significant improvement ($p < 0.01$) to all three approaches based on paired $t$-tests.

\begin{table}[!t]
\caption{Classification results for predicting between NP and P patients using the proposed method, compared to SVM, LLE and LL-LVM \cite{park2015bayesian} methods. Training performed using only shape information, only inter-vertebral (IV) poses and combined shape+IV poses as features. AUC=Area under curve.}
\label{table2}
\renewcommand{\arraystretch}{1.3}
\centering
\scalebox{0.95}{
\begin{tabular}{cccccc}
\hline
\hline
Data & Method & Accuracy  & Sensitivity & Specificity  & AUC\\
 \hline
Shape & SVM & 62.5 & 59.6 & 64.6 & 0.68\\
 & LLE & 66.2 & 63.5 & 67.1 & 0.70\\
 & LL-LVM & 70.7 & 72.7 & 67.9 & 0.76\\
 & \textbf{Proposed} & \textbf{75.5} & \textbf{80.4} & \textbf{73.2} & \textbf{0.79}\\
 \hline
Poses & SVM & 58.6 & 53.2 & 60.2 & 0.62\\
 & LLE & 60.8 & 59.0 & 64.7 & 0.65\\
 & LL-LVM & 69.0 & 70.7 & 71.6 & 0.74\\
 & \textbf{Proposed} & \textbf{77.9} & \textbf{80.1} & \textbf{77.2} & \textbf{0.81}\\
 \hline
  Shape + Poses & SVM & 63.5 & 58.3 & 65.1 & 0.69\\
  & LLE & 67.0 & 66.5 & 68.3 & 0.73\\
 & LL-LVM & 69.5 & 76.3 & 72.6 & 0.78\\
 & \textbf{Proposed} & \textbf{81.0} & \textbf{87.9} & \textbf{75.3} & \textbf{0.85}\\
\hline
\hline
\end{tabular}
}
\end{table}

\subsection{Spine Shape Prediction}

A clinical validation using patient data was conducted in order to assess the clinical accuracy of the predicted 3D reconstructions yielded by the proposed system. In this study, we compared the clinical and geometrical parameters between the actual longitudinal examinations and the generated 3D reconstructions based only on the input baseline model and bending information. The data used for the clinical study consisted of 40 adolescents with AIS from the databased described previously using a leave-one-out validation scheme, where their baseline spine reconstructions were obtained. Each case had two follow-up examinations prior to surgery, at 1 and 2 years. The inclusion criteria for this study was adolescent subjects who had their X-rays taken during a scoliosis clinic consultation for either diagnosis or follow-up, and a baseline angulation above $6^{\circ}$. This study group was comprised of 32 girls and 8 boys. The mean age of subjects was of $12 \pm 3$ (range 8-16) years old. The cohort in the study group was composed of only progressive patients.  The average main Cobb angle on the frontal plane at the first visit was $23^{\circ} \pm 7.9^{\circ}$. In addition to geometric comparisons between actual and predicted 3D reconstructions, a series of clinical 2D and 3D geometrical parameters were subsequently computed from these models and compared between both techniques. There were 7 right thoracic curves, 13 double curves (5 main thoracic, and 8 main left lumbar), 3 triple curves, 7 left thoracolumbar curves, and 10 either right lumbar or left thoracic curve.

\begin{figure}[tb]

  \centering
\begin{minipage}[b]{0.99\linewidth}
  \centering
  \includegraphics[height=1.7in] {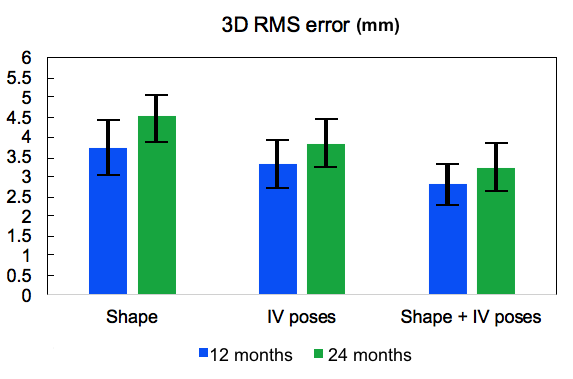}
\end{minipage}
\begin{minipage}[b]{0.99\linewidth}
  \centering
  \includegraphics[height=1.7in] {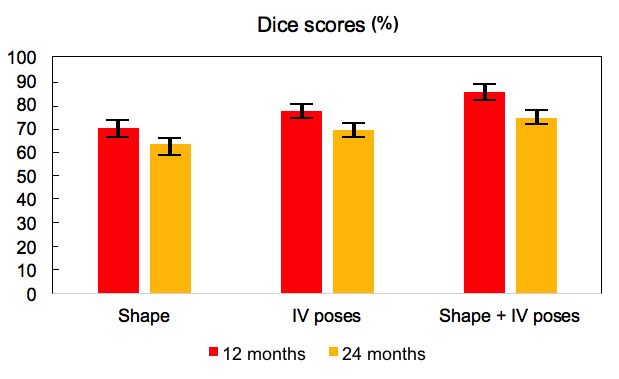}
\end{minipage}

\caption{Error bar plots for 3D root-mean-square difference of the vertebral landmarks generated and the Dice coefficients of the vertebral shapes.}
\label{fig:errorbarprediction}
\end{figure}

\begin{figure*}[tb]

  \centering
\begin{minipage}[b]{0.44\linewidth}
  \centering
  \includegraphics[height=1.8in] {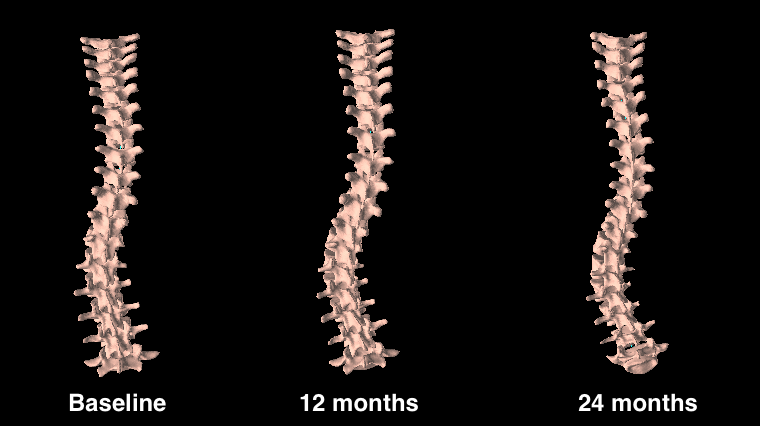}
        \centerline{(a)}
\end{minipage}
\begin{minipage}[b]{0.49\linewidth}
  \centering
  \includegraphics[height=1.8in] {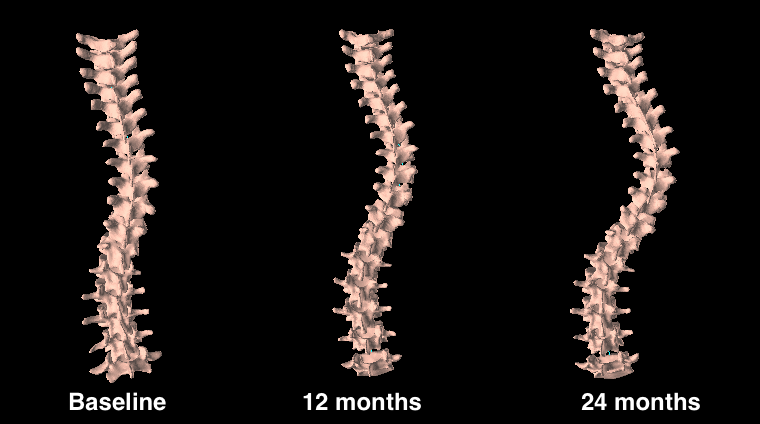}
        \centerline{(b)}
\end{minipage}
\begin{minipage}[b]{0.44\linewidth}
  \centering
  \includegraphics[height=1.8in] {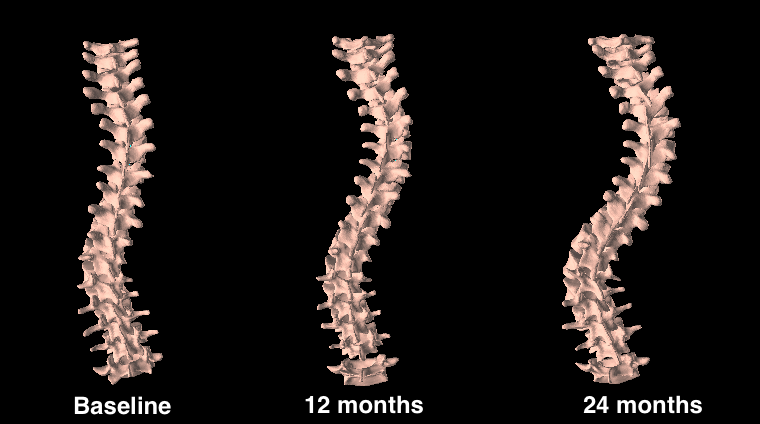}
        \centerline{(c)}
\end{minipage}
\begin{minipage}[b]{0.49\linewidth}
  \centering
  \includegraphics[height=1.8in] {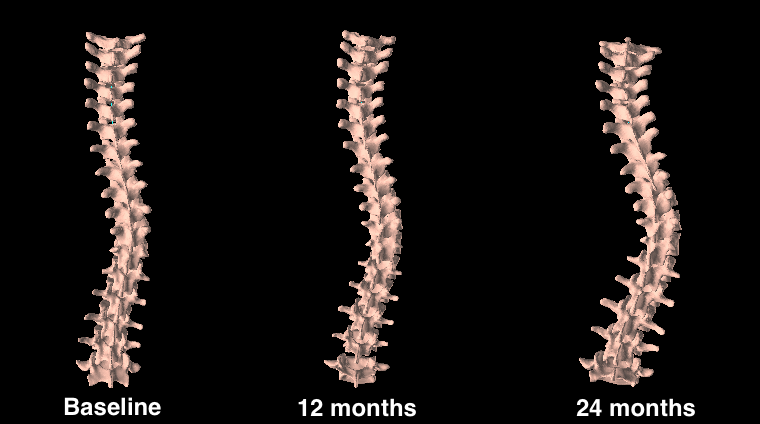}
        \centerline{(d)}
\end{minipage}

\caption{Examples of prediction results at 12 and 24 months based on the initial baseline 3D reconstruction. Patient cases are shown for (a) single curve left lumbar deformity, (b) major  right thoracic and minor left lumbar deformity, (a) double curve right thoracic left lumbar deformity and (b) single curve right thoracic deformity.}
\label{fig:exampleprediction}
\end{figure*}

We evaluated the geometrical accuracy of the predicted models at two time-points, at $t=12$ and $t=24$ months. Because the follow-up visits were not precisely, we grouped patients with a visit between 10 and 14 months within the 1-year bin, and visits between 22 and 26 months within the 2-year bin. For the predicted models, we evaluated both the 3D root-mean-square difference of the vertebral landmarks generated and the Dice coefficients of the vertebral shapes. The results are shown in Fig. \ref{fig:errorbarprediction}. We compared results using different composition for feature vectors $\textbf{y}$: 1) vertebral shape features, 2) inter-vertebral poses and 3) combination of shape+poses. Fig. \ref{fig:exampleprediction} shows sample prediction results at 12 and 24-months for 2 different clinical groups, which are commonly seen in the scoliotic population with thoracic and lumbar deformities. The results show encouraging predicted geometrical structures which offers a globally accurate representation of how the spine deformation has progressed at different structural levels. One can observe the local shape deformation is also well captured in the predicted models.

Table III presents the results from this clinical validation. The value of the computerized Cobb angle in the frontal plane ($C_{PA}^{PT}$, $C_{PA}^{MT}$, $C_{PA}^{L}$) and in the sagittal plane ($C_{LAT}^{T4-T12}$, $C_{LAT}^{L1-L5}$) are similar between ground-truth and predicted models with statistically insignificant differences ($p < 0.05$), while differences are slightly higher for 3D measurements such as the orientation of the planes of maximum curvature ($\theta_{PMC}^{PT}$, $\theta_{PMC}^{MT}$, $\theta_{PMC}^{L}$). Still, these differences remain  very acceptable, as they were found to be statistically insignificant ($p \leq 0.05$). Balance in the frontal and sagittal planes ($y_{T1-L5}$, $x_{T1-L5}$) on the other hand shows increased deviations based on known reconstructions, which can be attributed to the fact that the global position during acquisition is not a factor which is controlled or taken under consideration during the prediction phase. 


\begin{table}[!t]

\caption{RMS difference and Wilcoxon test results of the geometrical indices measured on ground-truth and predicted 3D reconstructions of the spine obtained from the proposed system at 24 months. NS: Non-significant difference. SD: Significant difference}
\label{table_3e}
\renewcommand{\arraystretch}{1.5}
\centering
\scalebox{0.95}{
\begin{tabular}{c c c c c}
\hline
\hline
Parameter & Symbol& Unit & Mean diff. & \emph{p}-value \\
\hline

Cobb angle (PT) & $C_{PA}^{PT}$ & deg & 2.5 $\pm$ 0.7 & 0.58 (NS)\\
\hline
Cobb angle (MT) & $C_{PA}^{MT}$ & deg & 2.1 $\pm$ 0.6 & 0.47 (NS)\\
\hline
Cobb angle (L) & $C_{PA}^{L}$ & deg & 2.8 $\pm$ 0.8 & 0.31 (NS)\\
\hline
Kyphosis & $C_{LAT}^{T4-T12}$ & deg & 3.2 $\pm$ 1.0 & 0.13 (NS)\\
\hline
Lordosis & $C_{LAT}^{L1-L5}$ & deg & 4.3 $\pm$ 1.3 & 0.19 (NS)\\
\hline
Max. deformity (PT) & $\theta_{PMC}^{PT}$ & deg & 4.6 $\pm$ 1.2 & 0.11 (NS)\\
\hline
Max. deformity (MT) & $\theta_{PMC}^{MT}$ & deg & 4.2 $\pm$ 0.9 & 0.17 (NS)\\
\hline
Max. deformity (L) & $\theta_{PMC}^{L}$ & deg & 4.9 $\pm$ 1.1 & 0.09 (NS)\\
\hline
Axial rotation & $\theta_{APEX}^{MT}$ & deg & 2.8 $\pm$ 0.7 & 0.33 (NS)\\
\hline
Frontal balance & $y_{T1-L5}$ & deg & 6.3 $\pm$ 2.2 & 0.02 (SD)\\
\hline
Sagittal balance & $x_{T1-L5}$ & deg & 8.7 $\pm$ 2.9 & 0.01 (SD)\\
\hline
\hline
\end{tabular}
}
\end{table}

\section{Discussion and conclusion}
\label{Conclusion}

We proposed a method to predict the progression of spinal deformities in patients diagnosed with AIS using geodesic parallel transport curves generated from probabilistic manifold models. Our main contribution consists in describing the time progression of complex spinal deformity patterns in a non-linear and discriminant Riemannian framework by first distinguishing  non-progressive and progressive cases, followed by a prediction of structural evolution. Articulated mesh models are described as a combination of both vertebral shape constellations and rigid inter-vertebral connections. Both high dimensional samples offer complementary features when learning the shape space of variations. To this end, we proposed a discriminant feature to the probabilistic model which links the low-dimensional manifold coordinates and the high-dimensional samples based on the rationale that functions are drawn from Gaussian priors. A particularity for discriminant manifolds with a probabilistic modeling of the inherent data structure is it ensures consistency within sub-regions where samples points which are less dense, thus creating smooth transitions in local neighborhoods of individual spatiotemporal trajectories describing curvature progression in the scoliotic population. This is achieved by representing graph relations with Gaussian priors which are drawn from similar distributions. Towards this end, a geodesic kernel is used to represent dissimilarities in object constellations present in the data, by measuring inter-vertebral variations as well as shape morphology.

The proposed model provides a way to analyse longitudinal samples from a geodesic curve in manifold space, thus simplifying the mixed effects when studying group-average trajectories. The model was used to predict progressive scoliosis diseases from a baseline reconstruction and skeletal parameters such as the bending flexibility ratio. We validated the spatiotemporal transformations for individual progression trajectories by comparing the predicted outcome from the parallel curves using the tangent plane exponential, to the actual 3D shape reconstruction. Three variables control the evolution: growth acceleration, time at baseline and shape morphology. Previous techniques for disease progression explicitly modeled using high-dimensional finite element models or from biomechanical simulations.

The results obtained from the prediction framework are concordant with a number of clinical findings, studying spinal deformity progression. Villemure et al \cite{villemure2001progression} found a concomitant progression between curve severity and 3D vertebral body wedging. This observation was also seen in this study as it reveals greater local vertebra deformation when time-shifts from the baseline reconstruction increases, thereby inducing increased vertebral angulation on 2-year predictions.
It is recognized from clinical experience that progression in AIS is primarily driven by skeletal and chronological age, as well as on the class of deformation (thoracic, lumbar), and the severity of the curve deformation. However these discrete parameters, such as curve magnitude obtained at the first visit are not sufficient to accurately predict whether the main curve will progress or not. The proposed framework is able to process the entire spine model which adds significant insight on the predominant features used for the prediction of curve progression. A previous  study by the Scoliosis Research Society 3D Scoliosis Committee demonstrated that similar 3D profiles can lead to different 3D morphology progression and thus stressed on quantifying 3D deformations. In \cite{nault2014three}, a number of clinical parameters including the angulation of the main curve and apical inter vertebral axial rotation were found to be leading predictors. The main problem in dividing the different geometries of deformation as potential risk factor of progression is the lack of robustness based on the accuracy of these measures.
Future directions of our research is to evaluate such framework in a prospective, multi-center study in order to study the affect of baseline 3D reconstruction quality, increase the size of the database to capture more classes of deformation, particularly more rare types which exhibit different progression behaviors than more typical deformations, such as single thoracic curves.

\ifCLASSOPTIONcaptionsoff
  \newpage
\fi


%

\bibliographystyle{IEEEtran}
\bibliography{strings,IEEEexample}

\end{document}